\documentclass[10pt,twocolumn,letterpaper]{article}

\usepackage[final,algorithms]{wacv}      

\usepackage{graphicx}
\usepackage{amsmath}
\usepackage{amssymb}
\usepackage{amsfonts}
\usepackage{booktabs}
\usepackage{algorithm}
\usepackage{algorithmic}
\usepackage{tikz}
\usetikzlibrary{positioning, arrows.meta, fit, shapes.geometric}
\usepackage{pgfplots}
\pgfplotsset{compat=1.17}
\usepackage{multirow}
\usepackage[table]{xcolor}
\usepackage{listings}
\usepackage{hhline}
\usepackage{subcaption}
\definecolor{voxelcolor}{rgb}{0.2, 0.5, 0.8} 
\definecolor{himcolor}{rgb}{0.8, 0.2, 0.2}   

\usepackage[pagebackref,breaklinks,colorlinks]{hyperref}

\def\wacvPaperID{3097} 
\def\confName{WACV}
\def\confYear{2026}

\begin{document}

\title{FocalComm: Hard Instance-Aware Multi-Agent Perception}

\author{Dereje Shenkut \quad Vijayakumar Bhagavatula\\
Carnegie Mellon University\\
{\tt\small \{dshenkut, vk16\}@andrew.cmu.edu}
}

\maketitle

\begin{abstract}
Multi-agent collaborative perception (CP) is a promising paradigm for 
improving autonomous driving safety, particularly for vulnerable road users like pedestrians, 
via robust 3D perception. However, existing CP approaches often optimize for vehicle detection performance metrics, underperforming on smaller, safety-critical objects such as pedestrians, where detection failures can be catastrophic.
Furthermore, previous CP methods rely on full feature exchange rather than 
communicating only salient features that help reduce false negatives. 
To this end, we present \textbf{FocalComm}, a novel collaborative perception framework that focuses
on exchanging hard-instance-oriented features among connected collaborative agents. 
FocalComm consists of two key novel designs, (1) a learnable progressive
hard instance mining (HIM) module to extract hard instances-oriented features per agent, and
(2) a query-based feature-level (intermediate) fusion technique that dynamically weights these 
identified features during collaboration. We show that FocalComm
outperforms state-of-the-art collaborative perception methods on two challenging real world datasets (V2X-Real and DAIR-V2X)
 across both vehicle-centric and infrastructure-centric collaborative setups. FocalComm also shows strong performance gain in
 pedestrian detection in V2X-Real. Code and model checkpoints are available at \url{https://github.com/scdrand23/FocalComm}.
\end{abstract}

\section{Introduction}
\label{sec:intro}

The ability to perceive and interpret the surroundings accurately is at the core of the promise of autonomous vehicle (AV) systems. 
Research in AVs \cite{Geiger2012CVPR, nuscenes, waymo2020} has made significant progress,
with the potential for a future of safer and more efficient intelligent transportation
systems. At the core of this progress lies robust perception. While single-vehicle perception has advanced significantly through multi-sensor fusion (cameras, LiDAR, radar) and data intensive learning-based techniques, it remains constrained by 
limited field-of-view, occlusions, and degraded performance at long ranges. Multi-agent\footnote{An agent refers to either a vehicle or infrastructure unit with both perception capability (e.g. LiDAR) and V2X communication modules that 
enable it to share and receive information.} CP where vehicles and infrastructure share
complementary perception data through V2X communication has shown promise to address this. CP enables connected autonomous vehicles (CAVs) and infrastructure units 
to exchange visual perception information and scene representations from multiple viewpoints, mitigating single-agent perception limitations and improving 
detection reliability for safety-critical scenarios. 

Collaboration can be at raw level (early fusion), feature level (intermediate fusion), or decision level (late fusion), 
based on the data sharing stage \cite{Han_2023}. Early fusion exchanges unprocessed sensor data (images, LiDAR point clouds),
requiring high communication bandwidth but enabling comprehensive joint processing. Late fusion transmits only final detection outputs (e.g., bounding boxes),
minimizing bandwidth needs but potentially introducing delays and information loss. Intermediate fusion strikes a balance by sharing 
compressed feature representations, offering a practical compromise between communication efficiency and perception performance. Current research has explored various aspects of CP, including bandwidth optimization and selective information transmission. 
Works such as Where2comm \cite{hu2022where2comm} have focused on optimizing bandwidth usage by selectively transmitting informative features, 
while others like V2X-ViT \cite{xu2022v2xvit} have addressed challenges such as noisy localization. Recent advances like SyncNet \cite{lei2022latencyaware} 
have made progress in latency-aware collaboration, while MPDA \cite{xu2023bridging} and DI-V2X \cite{DI_V2X_2023} have tackled domain gap in multi-agent collaboration. 

The detection of vulnerable road users such as pedestrians remains understudied in CP settings, both in terms of available datasets 
and methodological approaches. While current CP systems achieve good performance on vehicle detection, they significantly underperform on pedestrian 
detection. This performance gap is of much concern as pedestrian detection poses unique challenges due to their smaller size, occlusions, and weak sensing from long ranges, yet they 
represent critical safety risks when missed (nearly 7,522 pedestrian deaths in the USA in 2022 alone \cite{NHTSA_2024}). Therefore, there is a pressing need for CP systems that can effectively handle multi-class detection scenarios, 
particularly for safety-critical classes like pedestrians.

To address this challenge, we draw inspiration from a well-established line of work in single-agent perception: hard instance mining. These techniques, which identify and prioritize difficult-to-detect objects through loss-based sampling 
\cite{shrivastava2016training} or adaptive weighting \cite{lin2017focal, focalformer3d, xia2024hinted}, have shown significant promise.
They have proven particularly effective at improving detection performance on challenging objects by giving more focus on the most 
difficult examples during training, with recent works like FocalFormer3D \cite{focalformer3d} and HINTED \cite{xia2024hinted} demonstrating strong results on 
general 3D detection tasks. These approaches have led to improvements in detecting difficult instances such as small 
objects, partially occluded targets, 
and objects at long range. Motivated by these advances, we observe that hard instance focused collaboration could significantly benefit multi-agent perception by
 addressing communication constraints and detection challenges simultaneously. Adapting these techniques to CP requires careful architectural design to effectively prioritize difficult instances across multiple 
agents with  
bandwidth constraints. 

In this paper, we introduce FocalComm, a novel multi-agent CP method that prioritizes hard instance-oriented features based on a 
learnable difficulty-aware instance identification approach. Our approach consists of a stage-wise hard instance identification module followed by adaptive feature fusion module that 
selectively combines information from multiple agents based on instance-level difficulty queries. 
Our approach is motivated by the observation that not all 
 objects require equal collaborative effort. While vehicles are often detected reliably by a single agent, pedestrians and smaller objects benefit more from 
 multi-agent collaboration due to their size and likelihood of occlusion. By focusing exchanging information about these challenging instances, 
 FocalComm achieves superior performance on the real multi-class multi-agent collaborative perception benchmark \cite{v2xreal}. Our contributions in this paper can be summarized as follows.

\begin{enumerate}
\item We develop a multi-stage hard instance mining technique that extracts features ranked with detection uncertainty across multiple object classes from each agent to progressively focus on increasingly difficult detection cases.
\item We propose a query-guided multi-agent feature aggregation strategy that prioritizes hard instance-oriented queries across collaborative agents while dynamically weighting of features and queries from each agent.
\item We present FocalComm, an end-to-end framework that integrates difficulty-aware mining and query-guided fusion to achieve efficient CP, with particular benefits for hard-to-detect objects like pedestrians. 
\end{enumerate}

\section{Related Work}
\label{sec:related}
\input{sec/2_related_work}

\section{FocalComm Framework}
\label{sec:method}
\label{sec:methodology}
Our proposed FocalComm architecture is illustrated in Fig.~\ref{fig:focalcomm}. FocalComm processes inputs from multiple agents including the ego vehicle, other connected autonomous vehicles (CAVs), and infrastructure sensors.
Each agent's point cloud data ($X_e$ for ego, $X_i$ for the other CAVs, and $X_j$ for infrastructure) is first processed through identical sparse voxel feature encoders (denoted as $\Phi$), producing agent-specific feature maps ($F_e$, $F_i$, $F_j$). 
These features are then processed through two key components: (1) a progressive Hard Instance Mining (HIM) module that generates stage-wise heatmaps ($H_e^s$, $H_i^s$, $H_j^s$) and suppresses easy samples using mask ($\mathcal{M}$) to focus on more challenging objects across multiple stages, and
(2) Query-guided Adaptive Feature Fusion (QAFF) that aggregates the instance-aware queries from all agents into a unified representation, which is combined with concatenated BEV features ($F_{BEV}$) before feeding into the detection head.

\paragraph{Feature Extraction} Each agent processes its point cloud data through a shared sparse voxel feature encoder $\Phi$, producing agent-specific feature maps:
\begin{equation}
F_k = \Phi(X_k) \in \mathbb{R}^{H \times W \times C}, \quad k \in \{e, i, j\}
\end{equation}
where $X_k$ represents point cloud inputs from ego, CAV, and infrastructure agents.

\subsubsection{Hard Instance Mining (HIM)}
\begin{algorithm}[t]
       \caption{Hard Instance Mining (HIM)}
       \label{alg:phim}
       \begin{algorithmic}[1]
       \REQUIRE Multi-agent features $\mathcal{F}$, GT boxes $\mathcal{G}$, stages $S$
       \STATE $\mathcal{F}_{orig} \gets \mathcal{F}$ \hspace{2mm}
              $\triangleright$ Cache
       \STATE $\mathcal{M}_{acc} \gets \mathbf{0}$ \hspace{2mm}
              $\triangleright$ Initialize
       \FOR{$s \in \{1,\ldots,n_S\}$}
           \STATE $\mathcal{M}_{spatial} \gets \max(\mathcal{M}_{acc})$ \hspace{2mm}
                  $\triangleright$ Flatten
           \STATE $\mathcal{F}_{masked} \gets \mathcal{F}_{orig} \odot (1 - \mathcal{M}_{spatial})$ \hspace{2mm}
                  $\triangleright$ Mask
           \STATE $\hat{\mathcal{F}}_s \gets \Psi_s(\mathcal{F}_{masked})$ \hspace{2mm}
                  $\triangleright$ Extract
           \STATE $\mathcal{P}_s \gets \Omega(\hat{\mathcal{F}}_s)$ \hspace{2mm}
                  $\triangleright$ Detect
           \IF{training}
               \STATE $\mathcal{T}_s \gets \text{Match}(\mathcal{P}_s, \mathcal{G})$ \hspace{2mm}
                      $\triangleright$ Assign
           \ELSE
               \STATE $\mathcal{T}_s \gets \text{Filter}(\mathcal{P}_s)$ \hspace{2mm}
                      $\triangleright$ Threshold
           \ENDIF
           \STATE $\mathcal{M}_{acc} \gets \max(\mathcal{M}_{acc}, \mathcal{T}_s)$ \hspace{2mm}
                  $\triangleright$ Update
       \ENDFOR
       \STATE $\mathcal{Q} \gets \text{Combine}(\{\hat{\mathcal{F}}_s\})$ \hspace{2mm}
              $\triangleright$ Fuse
       \ENSURE Query features $\mathcal{Q}$, Predictions $\{\mathcal{P}_s\}$, Masks $\{\mathcal{T}_s\}$
       \end{algorithmic}
       \end{algorithm}
A key part of FocalComm is the Hard Instance Mining (HIM) module that extracts hard instance identifying features in a multi-stage manner. Each stage focuses on increasingly difficult instances while avoiding redundant attention to already-identified objects through mask accumulation.
As detailed in Algorithm~\ref{alg:phim}, HIM processes features through $n_S$ progressive stages. The key innovation lies in the progressive masking: at each stage $s$, the accumulated mask $\mathcal{M}_{acc}$ suppresses detected regions through $\mathcal{F}_{masked} = \mathcal{F}_{orig} \odot (1 - \mathcal{M}_{spatial})$, forcing subsequent stages to focus on harder instances.
The $\text{Match}()$ and $\text{Filter}()$ functions handle mask generation differently for training and inference. During training, $\text{Match}(\mathcal{P}_s, \mathcal{G})$ performs Hungarian assignment between predictions and ground truth, returning binary masks at locations where IoU exceeds $\tau_{iou}$. During inference, $\text{Filter}(\mathcal{P}_s)$ applies confidence thresholding: $\mathcal{T}_s = \mathbf{1}[\sigma(H_s) > \tau \cdot \gamma^s]$, where $H_s$ is the dense heatmap, $\tau$ is the base threshold, and $\gamma$ is a decay factor controlling stage-wise threshold progression.
Each stage produces features $\hat{\mathcal{F}}_s$ concatenated to form query features $\mathcal{Q} \in \mathbb{R}^{N \times (n_S \cdot C) \times H \times W}$.

\subsubsection{Query-guided Adaptive Feature Fusion (QAFF)}
\begin{algorithm}[t]
       \caption{Query-guided Adaptive Feature Fusion (QAFF)}
       \label{alg:qaff}
       \begin{algorithmic}[1]
       \REQUIRE Query features $\{\mathcal{Q}_s^i\}_{s=1}^{n_S}$ from each agent $i$, 
                Agent features $\mathcal{F}_i \in \mathbb{R}^{C \times H \times W}$,
                Valid agent mask $\mathbf{M} \in \{0,1\}^N$
       \FOR{$s \in \{1,\ldots,n_S\}$}
           \STATE $\tilde{\mathcal{Q}}_s \gets 
                  \text{MHSA}(\{\mathcal{Q}_s^i\}_i, \mathbf{M})$ \hspace{2mm}
                  $\triangleright$ Cross-agent attention
       \ENDFOR
       \STATE $\omega_s \gets \text{softmax}(\text{SA}(\tilde{\mathcal{Q}}_s))$ \hspace{2mm}
              $\triangleright$ Stage importance weights
       \STATE $\bar{\mathcal{Q}} \gets \sum_{s=1}^S \omega_s \tilde{\mathcal{Q}}_s$ \hspace{2mm}
              $\triangleright$ Stage-wise feature aggregation
       \STATE $\mathbf{K}, \mathbf{V} \gets \text{Proj}(\{\mathcal{F}_i\}_i)$ \hspace{2mm}
              $\triangleright$ Project features to key-value 
       \STATE $\mathcal{F}_{cross} \gets 
              \text{MHCA}(\bar{\mathcal{Q}}, \mathbf{K}, \mathbf{V}, \mathbf{M})$ \hspace{2mm}
              $\triangleright$ Query-guidance 
       \STATE $\alpha_i \gets \text{softmax}(\text{AA}(\mathcal{F}_{cross}) \odot \mathbf{M})$ \hspace{2mm}
              $\triangleright$ Agent weights
       \STATE $\mathcal{F}_{out} \gets \sum_i \alpha_i \mathcal{F}_{cross}^i$ \hspace{2mm}
              $\triangleright$ Weighted feature fusion
       \ENSURE Fused features $\mathcal{F}_{out} \in \mathbb{R}^{C \times H \times W}$
       \end{algorithmic}
       \end{algorithm}
The Query-guided Adaptive Feature Fusion (QAFF) module takes instance-aware queries from HIM to aggregate
information across multiple agents. As shown in Algorithm~\ref{alg:qaff}, QAFF takes as input the stage-wise query features
$\{\mathcal{Q}_s^i\}_{s=1}^S$ from each agent $i$, along with their original features $\mathcal{F}_i \in \mathbb{R}^{C \times
 H \times W}$ and a valid agent mask $\mathbf{M} \in \{0,1\}^N$ indicating participating agents in the scene.
 For stage $s$, QAFF first performs
 multi-head self-attention (MHSA) across agents to generate stage-specific representations $\tilde{\mathcal{Q}}_s$.
 This cross-agent attention mechanism allows agents to collaboratively refine their understanding of objects at each difficulty
 level while accounting for potentially missing or inactive agents through the mask $\mathbf{M}$. The stage-wise representations
 are then combined through learned importance weights $\omega_s$, computed via a stage attention (SA) mechanism and softmax normalization.
  This adaptive weighting scheme produces a unified query representation $\bar{\mathcal{Q}} = \sum_{s=1}^S \omega_s \tilde{\mathcal{Q}}_s$
 that emphasizes the most informative stages based on the current scene context. The original agent features $\{\mathcal{F}_i\}_i$ are
 projected into key-value space to obtain $\mathbf{K}$ and $\mathbf{V}$. Multi-head cross-attention (MHCA) is then applied between the
 unified queries $\bar{\mathcal{Q}}$ and these key-value pairs, producing $\mathcal{F}_{cross}$ that captures comprehensive multi-agent
 understanding. Finally, agent-specific attention weights $\alpha_i$ are computed through an agent attention (AA) mechanism, taking into
 account the valid agent mask $\mathbf{M}$. These weights determine each agent's contribution to the final fused output
 $\mathcal{F}_{out} = \sum_i \alpha_i \mathcal{F}_{cross}^i$, emphasizing agents with more informative observations. These weights automatically emphasize agents with more reliable or informative observations, accounting for variations in viewpoint quality and sensing capabilities.
The output of QAFF is then passed to the detection decoders (shown in purple in Fig.~\ref{fig:focalcomm}) for final object detection and classification.

\subsection{Detection Decoder and Joint Optimization}
The final fused features, $F_{Fuse}$, are passed to the detection head. Different from previous 
multi-agent collaborative perception works, we adopt an anchor-free detection head 
\cite{bai2021transfusion} that supervises the regression and classification tasks as well as 
allows joint optimization of the multistage hard instance identification. This anchor-free approach eliminates the need for complex 
predefined anchor designs and provides more direct object localization, which is particularly beneficial for detecting hard instances
with unusual scales or occlusion patterns that traditional anchor-based methods might struggle with. Furthermore, our approach naturally
handles multi-class detection scenarios, effectively identifying various road users including pedestrians, which 
exhibit significant variation in size and appearance. Our detection 
head consists of a transformer decoder based on \cite{bai2021transfusion} that processes feature 
queries from a dense heatmap prediction branch. The detection tasks are jointly optimized with hierarchical 
instance mining through a multi-component loss function:
\begin{equation}
\label{eq:loss}
\mathcal{L} = \lambda_{1}\mathcal{L}_{cls} + \lambda_{2}\mathcal{L}_{bbox} + 
\lambda_{3}\mathcal{L}_{hm} + \lambda_{4}\sum_{s=1}^{S}\mathcal{L}_{him}^{s}
\end{equation}
where $\mathcal{L}_{cls}$ is the focal classification loss, $\mathcal{L}_{bbox}$ is the L1 regression loss for bounding box parameters, $\mathcal{L}_{hm}$ is the Gaussian focal loss for heatmap prediction.
$\mathcal{L}_{him}^{s}$ represents the progressive loss for the multi-stage hard instance mining at stage $s$. 
The weights $\lambda_{1}$, $\lambda_{2}$, $\lambda_{3}$, and $\lambda_{4}$ balance the contributions of different loss components.
The specific values of these hyperparameters are provided in the implementation details section.

\section{Experiments}
\label{sec:experiments}
\begin{table*}
  \begin{center}
  \scriptsize
  \setlength{\tabcolsep}{3pt}
  \begin{tabular}{l|c|c|c|c|c|c|c|c|c|c}
  \hline
  \multirow{4}{*}{Method} & \multicolumn{8}{c|}{\textbf{V2XReal ($K_{max} = 4$)}} & \multicolumn{2}{c}{\textbf{DAIR-V2X ($K_{max} = 2$)}} \\
  \hhline{~|*{10}{-}|}
  & \multicolumn{2}{c|}{Car} & \multicolumn{2}{c|}{Pedestrian} & \multicolumn{2}{c|}{Truck} & \multicolumn{2}{c|}{Overall} & \multicolumn{2}{c}{Vehicle} \\
  \hhline{~|*{8}{-}|*{2}{-}|}
  & VC & IC & VC & IC & VC & IC & VC & IC & \multicolumn{2}{c}{VC}\\
  \hhline{~|*{8}{-}|*{2}{-}|}
  & AP@0.3/0.5 & AP@0.3/0.5 & AP@0.3/0.5 & AP@0.3/0.5 & AP@0.3/0.5 & AP@0.3/0.5 & mAP@0.3/0.5 & mAP@0.3/0.5 & {AP@0.3} & {AP@0.5} \\
  \hline\hline
  
  No Collaboration & 73.7/68.4 & 70.6/59.1 & 31.8/13.9 & 29.7/10.7 & 21.2/15.7 & 46.6/42.0 & 42.2/32.7 & 49.0/37.3 & 58.9 & 54.4 \\

  F-Cooper \cite{fcooper} & 88.3/85.6 & 84.3/80.8 & 47.8/22.7 & 45.4/15.9 & 47.9/46.1 & 48.3/47.9 & 61.3/51.4 & 59.4/48.2 & 70.4 & 64.8 \\
  V2VNet \cite{wang2020v2vnet} & 87.0/84.4 & 85.0/81.4 & 34.5/13.9 & 36.5/15.2 & 40.0/36.8 & 44.3/41.9 & 53.8/45.0 & 55.3/46.2 & 69.5 & 63.5 \\
  Attfuse \cite{xu2023bridging} & 81.3/80.7 & 81.5/80.9 & 46.8/21.7 & 48.5/24.8 & 49.6/47.7 & 47.6/45.7 & 59.2/50.0 & 59.2/50.5 & 69.7 & 63.8 \\
  CoBEVT \cite{hu2022where2comm} & 87.2/85.6 & 84.1/82.1 & 54.8/26.1 & \textbf{52.3}/25.6 & 50.1/45.1 & 48.9/47.8 & 64.0/53.3 & 61.7/\textbf{52.9} & 72.8 & 65.7 \\
  V2XViT \cite{xu2022v2xvit} & 83.9/81.1 & 81.4/78.2 & 38.5/15.2 & 33.5/13.3 & 42.5/35.6 & 45.4/38.9 & 55.0/44.0 & 53.4/43.5 & 74.5 & 67.6 \\
  CoAlign \cite{lu2023robust} & 85.8/83.4 & 84.7/83.4 & 38.3/17.3 & 36.4/14.8 & 52.7/43.9 & \textbf{53.2/51.1} & 59.9/48.2 & 58.1/49.8 & 76.9 & 69.7 \\
  ERMVP \cite{Zhang_2024_CVPR} & 88.5/86.4 & \textbf{86.7}/84.0 & 53.2/25.4 & 50.6/23.5 & 42.9/41.3 & 41.7/38.7 & 61.5/51.0 & 59.7/48.7 & 69.2 & 63.4 \\



  FocalComm (ours) & \textbf{91.5/89.6} & 86.2/\textbf{84.8} & \textbf{57.4/27.3} & 51.2/\textbf{26.7} & \textbf{53.9/51.6} & 49.6/47.3 & \textbf{67.6/56.1} & \textbf{62.3}/\textbf{52.9} & \textbf{77.2} & \textbf{70.1} \\

  \hline
  \end{tabular}
  
  \end{center}
  \caption{Performance comparison on V2X-Real dataset under Vehicle-Centric (VC) and Infrastructure-Centric (IC) collaborative setups. Results show AP@0.3/AP@0.5 format. Best results are in boldface. In no collaboration mode, VC means vehicle only and IC means infrastructure only. $K_{\max}$ is the maximum number of agents per scene.}
  \label{tab:v2xreal_results}
  \end{table*}
We evaluate \textbf{FocalComm} on V2XReal \cite{v2xreal} and DAIR-V2X \cite{yu2022dairv2x} datasets. To the best of our knowledge, V2X-Real  is  the only publicly accessible multi-agent collaborative real
dataset with enough multiclass annotation including pedestrians. For training, we randomly assign one agent as the ego vehicle, while at inference time we use predefined ego agents based on the dataset's categorization. 
Our experiments evaluate performance under two configurations,\textit{ i.e.}
vehicle-centric and infrastructure-centric, where the ego agent is vehicle and infrastructure, respectively.
For a fair comparison, all methods are implemented using the same 3D backbone \cite{yan2018second} and
anchor-free head \cite{bai2021transfusion}. We adopt a voxel-based method and anchor-free detection head across all compared models to push for improved detection performance for smaller classes such as pedestrians while ensuring fair comparison in collaborative perception.

\subsection{Datasets}
\paragraph{V2X-Real.} V2X-Real \cite{v2xreal} is a large-scale real-world dataset designed for V2X cooperative perception.  It includes 33K LiDAR frames and over 1.2 million annotated 3D bounding boxes. The dataset is collected using two connected automated vehicles and two smart infrastructures.  The dataset is collected in two scenarios: V2X smart intersections and V2V corridors. There are a maximum of four agents in a scene. The V2X-Real dataset contains multi-class (vehicle, pedestrian, and truck ) 
annotation and it allows vehicle-centric and infrastructure-centric CP 
evaluation. We use the train/val/test split with $23379$, $2770$, and $6850$ 
frames respectively as proposed in the benchmark \cite{v2xreal}.


\paragraph{DAIR-V2X} DAIR-V2X \cite{yu2022dairv2x} is the first large-scale real-world dataset for Vehicle-Infrastructure cooperative perception. The dataset comprises 71K LiDAR and camera frames collected from real scenarios with comprehensive 3D annotations. It features vehicle-infrastructure collaboration with temporal asynchrony challenges and includes V2X-Seq extension with 15K frames for sequential perception and trajectory forecasting tasks.

\paragraph{Implementation Details.}
During training, we voxelize the point cloud with a voxel size of $0.2m \times 0.2m \times 0.4m $ 
and use a range of $[-100m, 100m]$ for the $x$ and $y$ axes, and $[-10m, 6m]$ for the $z$ axis. 
Each voxel aggregates up to $20$ points. We implement FocalComm using PyTorch and train on four H100 GPUs with a batch size of $8$ for $50$ epochs
with Adam \cite{kingma2014adam} optimizer with learning rate starting at $1e^{-4}$ and applying a weight decay of $1e^{-2}$. 
Our model employs a standard sparse 3D CNN backbone commonly used in LiDAR-based 3D detection,
similar to VoxelNet \cite{zhou2018voxelnet} and SECOND \cite{yan2018second}. The progressive hard instance mining (HIM) module employs a multi-stage architecture with confidence 
thresholds of $0.4$, utilizing a pooling kernel size of $3$ for local detection peaks and masking 
with an attention decay factor of $2.0$ to progressively identify challenging instances.
Our query-guided adaptive feature fusion (QAFF) module employs multi-head attention with $8$ heads 
and hidden dimension of $256$ to dynamically fuse features across agents based on query importance.
The detection head follows a TransFusion \cite{bai2021transfusion} head with separate prediction 
branches for center, height, dimension, and rotation. Our multi-component loss function balances detection and mining objectives with weights
$\lambda_1\!=\!1.0$ for classification, $\lambda_2\!=\!2.0$ for bounding box regression, $\lambda_3\!=\!1.0$ for heatmap prediction,
and $\lambda_4\!=\!0.5$ for the hard instance mining component per agent. These weights were determined through extensive
ablation studies on the validation set. The weights $\lambda_1$-$\lambda_3$
follow established practices in prior detection works \cite{bai2021transfusion}, while $\lambda_4$
was specifically tuned for our approach to prevent overfitting to difficult examples early
in training while still ensuring sufficient gradient flow for learning challenging cases. This balance ensures stable
convergence while maintaining focus on both common and rare detection scenarios.

\subsection{Quantitative Evaluation}
\paragraph{Evaluation Protocol.} 
Similar to the evaluation protocol in V2VNet \cite{wang2020v2vnet} and V2X-Real \cite{v2xreal}, our evaluation is done in the range of $[-100m, 100m]$ 
in both the $x-axis$ and $y-axis$ of the chosen ego agent. We adopt the standard average precision (AP) at intersection-over-union (IoU) 
threshold of $0.3$ and $0.5$ for each class and mean average precision averaged 
overall number of classes. Following established practices for datasets in V2X-Real
with significant object size variations, we use these lower IoU thresholds to account for the 
challenging nature of detecting objects ranging from small pedestrians to large vehicles in CP scenarios.
\paragraph{Performance Comparison.}
Table~\ref{tab:v2xreal_results} presents the comparison between our FocalComm and existing methods on the V2X-Real \cite{v2xreal} dataset
under both infrastructure-centric (IC) and vehicle-centric (VC) settings. Our method achieves state-of-the-art performance across all metrics,
with significant improvements. In the vehicle-centric setting, FocalComm achieves 67.6\% mAP@0.3 and 56.1\% mAP@0.5, representing a 5.6\% and 5.1\% absolute improvement over the next best method (CoBEVT). The performance gains are particularly pronounced for pedestrian detection, where FocalComm achieves 57.4\% AP@0.3 and 27.3\% AP@0.5 in the vehicle-centric setting, significantly outperforming all baselines and addressing a critical safety need. Similarly, truck detection improves substantially from 21.2\% to 53.9\% AP@0.3, demonstrating our method's effectiveness on challenging large objects. For infrastructure-centric scenarios, our method maintains strong performance with 62.3\% mAP@0.3 and 52.9\% mAP@0.5. On the DAIR-V2X dataset for vehicle detection, FocalComm achieves competitive results with 77.2\% AP@0.3 and 70.1\% AP@0.5, demonstrating strong generalization across different V2X scenarios.

Table~\ref{tab:v2v_i2i_results} further validates our approach in specific communication scenarios. In V2V settings, FocalComm reaches 64.8\% mAP@0.3, while in I2I configurations it achieves an impressive 70.2\% mAP@0.3 and 58.1\% mAP@0.5, outperforming all baselines by substantial margins. Analyzing per-class patterns, I2I consistently outperforms V2V: truck detection benefits most (+10.3\% AP@0.3, from 50.5\% to 60.8\%) due to infrastructure's elevated viewpoint providing better coverage of large objects, while pedestrian detection gains +4.6\% (53.9\% to 58.5\%) from reduced occlusions. Vehicle detection shows a smaller gap (+1.3\%) as cars are well-detected from either perspective. These results demonstrate our method's ability to effectively leverage the complementary viewpoints available in different collaboration scenarios.

\subsection{Qualitative Evaluation}
\begin{figure*}[t]
  \centering
  \includegraphics[width=1.0\linewidth, height=0.3\textheight, keepaspectratio]{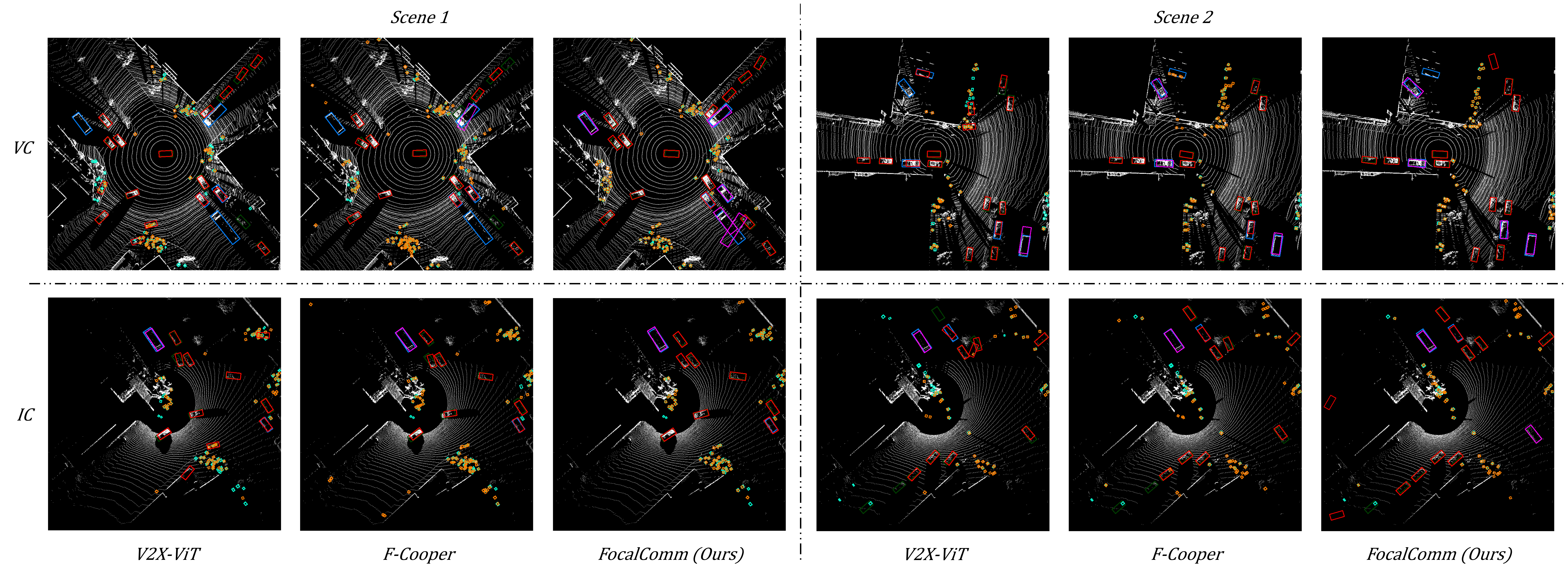}
  \caption{Qualitative detection results of FocalComm, V2XViT~\cite{xu2022v2xvit}, and F-Cooper~\cite{fcooper} on V2X-Real (cropped to $80\!\times\!80$m around ego). Ground truth and predictions are shown for Car (GT: \textcolor[RGB]{0,65,0}{dark green}, Pred: \textcolor[RGB]{255,0,0}{red}), Pedestrian (GT: \textcolor[RGB]{0,255,210}{cyan}, Pred: \textcolor[RGB]{255,128,0}{orange}), and Truck (GT: \textcolor[RGB]{0,128,255}{light blue}, Pred: \textcolor[RGB]{255,0,255}{magenta}).}
  \label{fig:qualitative}
\end{figure*}

\paragraph{Detection Results.}
We visualize detection results from both infrastructure-centric (IC) and vehicle-centric (VC) perspectives across two scenes in Figure~\ref{fig:qualitative}. 
Scene 1 (left) and Scene 2 (right) demonstrates our method's performance at a complex intersection from bird's-eye view. In Scene 1's dense intersection, FocalComm accurately detects 
crowded pedestrians and most of the trucks, where previous methods often struggle with occlusions and object overlap. 
Scene 2 highlights our method's effectiveness in detecting distant objects and multiple object classes. 
\begin{table*}
  \begin{center}
  \scriptsize
  \setlength{\tabcolsep}{2.5pt}
  \begin{tabular}{l|c|c|c|c|c|c|c|c}
  \hline
  \multirow{4}{*}{Method} & \multicolumn{8}{c}{\textbf{V2X-Real Communication Scenarios}} \\
  \hhline{~|*{8}{-}|}
  & \multicolumn{2}{c|}{Vehicle} & \multicolumn{2}{c|}{Pedestrian} & \multicolumn{2}{c|}{Truck} & \multicolumn{2}{c}{Overall} \\
  \hhline{~|*{8}{-}|}
  & V2V & I2I & V2V & I2I & V2V & I2I & V2V & I2I \\
  \hhline{~|*{8}{-}|}
  & AP@0.3/0.5 & AP@0.3/0.5 & AP@0.3/0.5 & AP@0.3/0.5 & AP@0.3/0.5 & AP@0.3/0.5 & mAP@0.3/0.5 & mAP@0.3/0.5 \\
  \hline\hline
  
  No Collaboration & 73.7/68.4 & 70.6/59.1 & 31.8/13.9 & 29.7/10.7 & 21.2/15.7 & 46.6/42.0 & 42.2/32.7 & 49.0/37.3 \\
  F-Cooper \cite{fcooper} & 86.6/83.2 & 84.0/79.9 & 45.3/23.3 & 49.7/21.0 & 45.5/40.8 & 59.3/58.2 & 59.1/49.1 & 64.3/53.0 \\
  V2VNet \cite{wang2020v2vnet} & 86.5/82.5 & 86.7/82.1 & 31.1/13.1 & 41.9/18.9 & 39.2/32.9 & 53.5/49.8 & 52.3/42.8 & 60.7/50.3 \\
  AttFuse \cite{xu2023bridging} & 81.1/79.9 & 82.8/81.8 & 44.4/20.5 & 52.1/28.2 & 48.7/\textbf{46.5} & 57.4/55.3 & 58.1/49.0 & 64.1/55.1 \\
  CoBEVT \cite{hu2022where2comm} & 86.1/83.9 & 84.0/81.1 & 51.0/\textbf{28.5} & 53.1/\textbf{30.6} & 48.6/43.3 & \textbf{61.9/60.1} & 61.9/51.9 & 66.3/57.2 \\
  V2X-ViT \cite{xu2022v2xvit} & 84.1/80.6 & 84.5/80.2 & 38.2/15.6 & 38.7/15.7 & 41.4/37.0 & 54.6/53.4 & 54.5/44.4 & 59.3/49.7 \\
  CoAlign \cite{lu2023robust} & 83.6/80.8 & 83.5/82.0 & 37.4/17.1 & 41.1/17.4 & 50.1/36.1 & 57.0/54.4 & 57.0/44.7 & 60.5/51.3 \\
  ERMVP \cite{Zhang_2024_CVPR} & 86.7/84.0 & 84.7/82.0 & 50.6/23.5 & 52.2/27.2 & 41.7/38.7 & 55.7/55.0 & 59.7/48.7 & 64.2/54.7 \\
  \hline
  FocalComm (ours) & \textbf{90.0/87.0} & \textbf{91.3/87.5} & \textbf{53.9}/27.1 & \textbf{58.5}/29.5 & \textbf{50.5}/42.3 & 60.8/57.4 & \textbf{64.8/52.1} & \textbf{70.2/58.1} \\
  \hline
  \end{tabular}
  \end{center}
  \caption{Performance comparison on V2X-Real dataset for Vehicle-to-Vehicle (V2V) and Infrastructure-to-Infrastructure (I2I) communication scenarios. Results show AP@0.3/AP@0.5 format. For FocalComm, I2I corresponds to our normal collaborative setup. Best results are in boldface.}
  \label{tab:v2v_i2i_results}
\end{table*}
\begin{figure*}[t]
  \centering
  \includegraphics[width=0.9\linewidth]{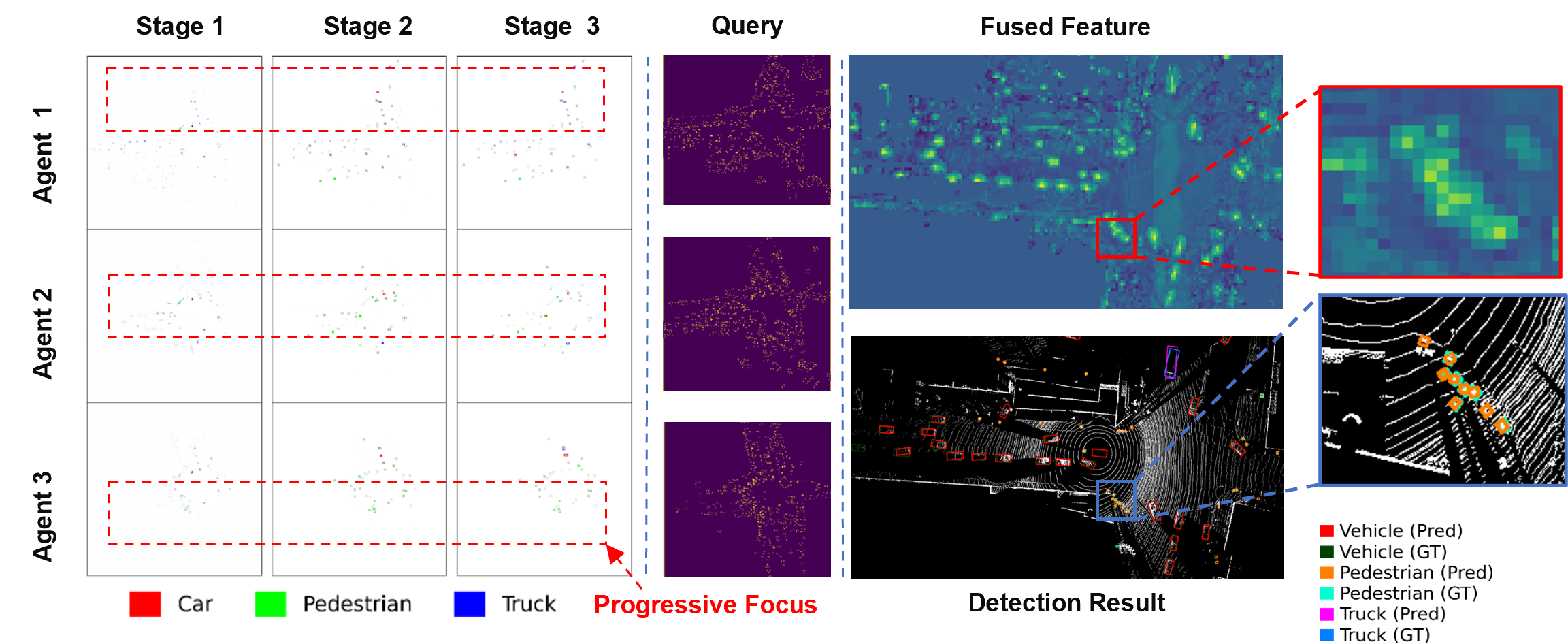}
  \caption{Visualization of our multi-stage Hard Instance Mining (HIM) process and query feature generation.
  The figure shows results from two agents (rows) across three HIM stages (first three columns), with detections color-coded as Car
   (\textcolor[RGB]{0,255,0}{green}), Pedestrian (\textcolor[RGB]{255,0,255}{magenta}), and Truck (\textcolor[RGB]{0,0,255}{blue}).
   The fourth column shows query feature heatmaps where brighter colors indicate higher attention weights.
    The rightmost column provides detection results. This progression demonstrates how our model systematically refines detection
     focus across stages in a multi-agent setup.}
  \label{fig:qaff_vis}
\end{figure*}
Through effective multi-agent collaboration, our method successfully detects cars (shown in \textcolor[RGB]{255,0,0}{red}), pedestrians (shown in \textcolor[RGB]{255,128,0}{orange}), and trucks (shown in \textcolor[RGB]{255,0,255}{magenta}) 
across varying distances. The method particularly excels at maintaining reliable detection performance for distant objects and handling cases 
where objects are only partially visible. Our method achieves high precision with very few false positives, while maintaining high recall with minimal false negatives,
as evidenced by the close alignment between predicted boxes and ground truth annotations across both scenes.
\paragraph{Query Features.}
To provide insights into our model's attention mechanism, we visualize the mean of query features and their progression across the three stages
of Hard Instance Mining (HIM) in Figure~\ref{fig:qaff_vis}. The visualization shows results from two different agents, with each row representing 
an agent's perspective. The first three columns show the evolution of instance detection across stages, using color coding to distinguish between
cars (\textcolor[RGB]{0,255,0}{green}), pedestrians (\textcolor[RGB]{255,0,255}{magenta}), and trucks (\textcolor[RGB]{0,0,255}{blue}). We observe that Stage 1 captures a broader set of potential instances, while Stages 2 and 3 
progressively refine and focus on previously undetected cases. The fourth column displays the generated query features as heatmaps along detection output, 
where brighter regions indicate higher weights. 

\subsection{Ablation Study}
\paragraph{Core Component Analysis.}
To validate the effectiveness of our key components, we conduct ablation studies showing the incremental contribution of each module.
As shown in Table~\ref{tab:ablation}, starting from a no-collaboration baseline with 42.2\% overall AP@0.3, basic collaborative
fusion (F-Cooper) improves performance to 61.3\% (+19.1\% absolute). Adding HIM alone to collaboration yields 66.2\% AP@0.3 (+4.9\% over baseline collaboration), while adding QAFF alone achieves
65.5\% AP@0.3 (+4.2\% over baseline). The full model combining both HIM and QAFF reaches 67.6\% AP@0.3, demonstrating synergy between the
components with an additional 1.4-2.1\% gain over individual components. The improvements are particularly notable for pedestrian detection (from 31.8\% to 57.4\% AP@0.3) and truck detection (from 21.2\% to 53.9\%),
validating our focus on hard instance mining for challenging objects.

\begin{figure}[t]
  \centering
  \includegraphics[width=\linewidth]{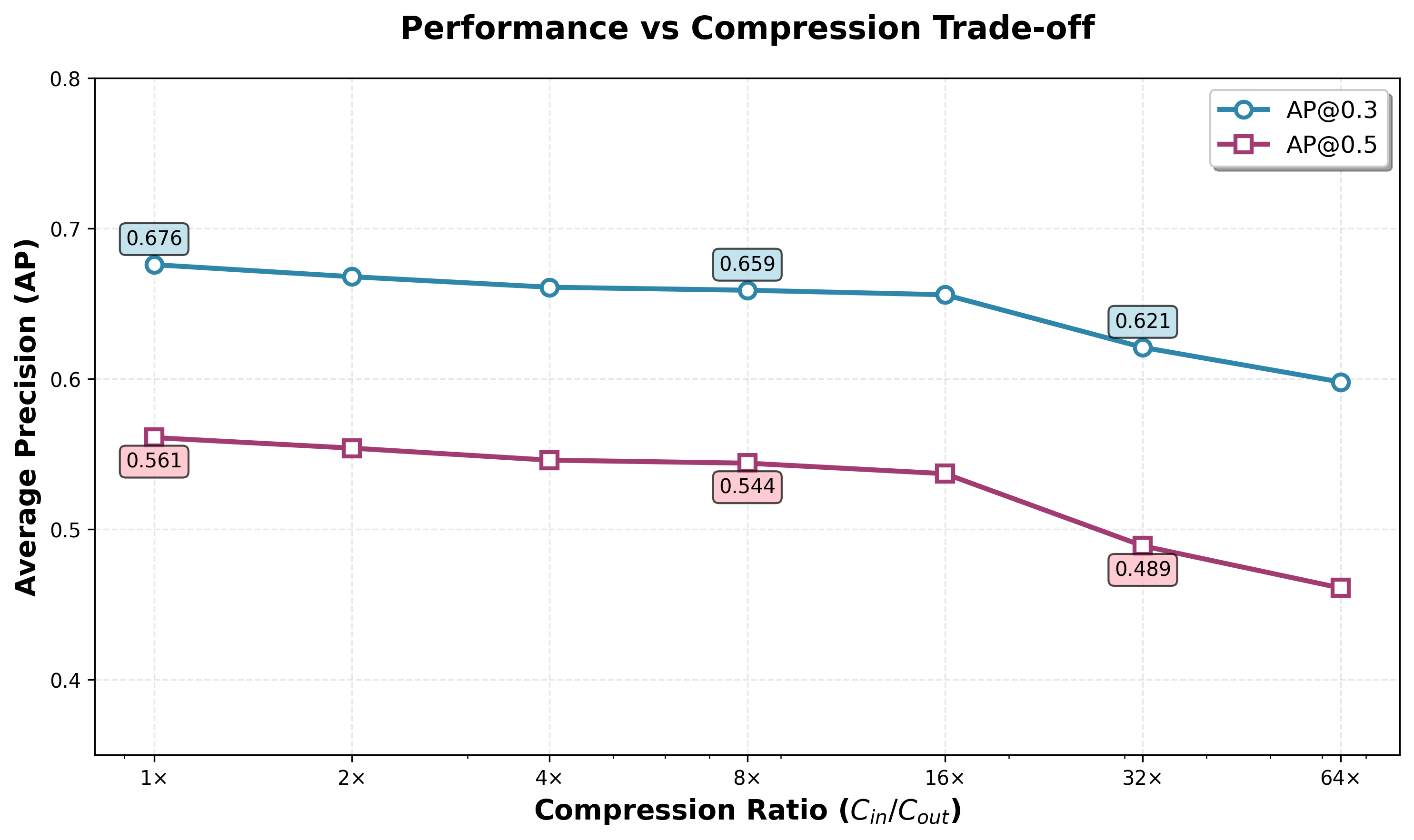}
  \caption{Performance vs compression trade-off analysis on V2X-Real dataset. FocalComm maintains robust performance up to 8× compression with minimal degradation, demonstrating effective communication efficiency for practical V2X deployment.}
  \label{fig:compression_vs_ap}
\end{figure}

\paragraph{Compression Analysis.}
To evaluate the communication efficiency of FocalComm, we analyze the trade-off between detection performance and feature compression ratios. Figure~\ref{fig:compression_vs_ap} shows how our method's performance varies across different compression levels from 1× (no compression) to 64× compression. We observe that FocalComm maintains robust performance up to 8× compression, with only a 2.5\% drop in AP@0.3 (from 67.6\% to 65.9\%) and minimal degradation in AP@0.5 (from 56.1\% to 54.4\%). Even at aggressive 32× compression, the method retains 62.1\% AP@0.3 and 48.9\% AP@0.5, demonstrating the effectiveness of our hard instance-focused feature selection. This analysis validates that FocalComm can operate efficiently under bandwidth constraints while maintaining strong detection performance, making it practical for real-world V2X deployment scenarios.

\begin{table}[t]
  \begin{center}
\scriptsize
\setlength{\tabcolsep}{5pt}
  \begin{tabular}{l|c|c|c|c}
  \hline
  Method & Car & Pedestrian & Truck & Overall \\
  \hline
  \hline
  No Collaboration & 73.7/68.4 & 31.8/13.9 & 21.2/15.7 & 42.2/32.7 \\
  \hline
  \multicolumn{5}{l}{\textit{Component Analysis}} \\
  \hline
  + Collab (baseline) & 88.3/85.6 & 47.8/22.7 & 47.9/46.1 & 61.3/51.4 \\
  + Collab + HIM & 91.5/88.8 & 54.6/26.6 & 52.6/48.7 & 66.2/54.7 \\
  + Collab + QAFF & 91.2/88.2 & 52.7/23.7 & 48.7/45.1 & 65.5/54.0 \\
  \hline
  Full Model (+ Both) & \textbf{91.5/89.6} & \textbf{57.4/27.3} & \textbf{53.9/51.6} & \textbf{67.6/56.1} \\
  \hline
  \end{tabular}
  \end{center}
  \caption{Ablation studies on V2X-Real test set for the vehicle-centric approach.
  Results show AP@0.3/AP@0.5 for each class.}
  \label{tab:ablation}
\end{table}


\section{Conclusion}
\label{sec:conclusion}
We presented FocalComm, a novel multi-agent collaborative perception framework that prioritizes hard instance-aware feature exchange. Our progressive HIM module and QAFF mechanism achieve state-of-the-art results 67.6\% mAP@0.3 on V2X-Real (5.6\% improvement) with strong performance in both V2V (64.8\%) and I2I (70.2\%) scenarios. FocalComm excels at safety-critical pedestrian detection (57.4\% AP@0.3, 80\% relative improvement) while maintaining robust performance under 8× compression.

\noindent\textbf{Limitations.} Our evaluation focuses on LiDAR-based datasets; extending to camera-only or multimodal fusion remains future work, along with formal theoretical analysis of HIM convergence and evaluation under adverse weather conditions.
Future work will explore hard-instance aware message packing, multimodal extensions, and theoretical grounding for real-world V2X deployment. 

\noindent\textbf{Acknowledgment.} This work was supported by US DOT Safety 21 University Transportation Center, Carnegie Mellon University, Pittsburgh, PA, USA.



{\small
\begin{thebibliography}{10}

\bibitem{bai2021transfusion}
Xuyang Bai, Zeyu Hu, Xinge Zhu, Qingqiu Huang, Yilun Chen, Hongbo Fu, and Chiew-Lan Tai.
\newblock {TransFusion}: Robust {LiDAR}-camera fusion for {3D} object detection with transformers.
\newblock In {\em Proceedings of the IEEE/CVF Conference on Computer Vision and Pattern Recognition}. IEEE, 2022.

\bibitem{nuscenes}
Holger Caesar, Varun Bankiti, Alex~H. Lang, Sourabh Vora, Venice~Erin Liong, Qiang Xu, Anush Krishnan, Yu~Pan, Giancarlo Baldan, and Oscar Beijbom.
\newblock nu{S}cenes: A multimodal dataset for autonomous driving.
\newblock In {\em Proceedings of the IEEE/CVF Conference on Computer Vision and Pattern Recognition}, pages 11618--11628, Seattle, WA, 2020. IEEE.

\bibitem{cai2018cascade}
Zhaowei Cai and Nuno Vasconcelos.
\newblock Cascade {R-CNN}: Delving into high quality object detection.
\newblock In {\em Proceedings of the IEEE Conference on Computer Vision and Pattern Recognition}, pages 6154--6162. IEEE, 2018.

\bibitem{fcooper}
Qi~Chen, Xu~Ma, Sihai Tang, Jingda Guo, Qing Yang, and Song Fu.
\newblock {F-Cooper}: Feature based cooperative perception for autonomous vehicle edge computing system using {3D} point clouds.
\newblock In {\em Proceedings of the IEEE/ACM Symposium on Edge Computing}, pages 88--100. IEEE, 2019.

\bibitem{chen2022co3}
Runjian Chen, Yao Mu, Runsen Xu, Wenqi Shao, Chenhan Jiang, Hang Xu, Zhenguo Li, and Ping Luo.
\newblock {CO3}: Cooperative unsupervised {3D} representation learning for autonomous driving.
\newblock In {\em Proceedings of the International Conference on Learning Representations}. ICLR, 2023.

\bibitem{focalformer3d}
Yilun Chen, Zhiding Yu, Yukang Chen, Shiyi Lan, Anima Anandkumar, Jiaya Jia, and Jose~M Alvarez.
\newblock {FocalFormer3D}: Focusing on hard instance for {3D} object detection.
\newblock In {\em Proceedings of the IEEE International Conference on Computer Vision}. IEEE, 2023.

\bibitem{cui2022coopernaut}
Jiaxun Cui, Hang Qiu, Dian Chen, Peter Stone, and Yuke Zhu.
\newblock Coopernaut: End-to-end driving with cooperative perception for networked vehicles.
\newblock In {\em Proceedings of the IEEE/CVF Conference on Computer Vision and Pattern Recognition}, pages 17252--17262. IEEE, 2022.

\bibitem{sst2022}
Lue Fan, Ziqi Pang, Tianyuan Zhang, Yu-Xiong Wang, Hang Zhao, Feng Wang, Naiyan Wang, and Zhaoxiang Zhang.
\newblock Embracing single stride {3D} object detector with sparse transformer.
\newblock In {\em Proceedings of the IEEE/CVF Conference on Computer Vision and Pattern Recognition}, pages 8458--8468. IEEE, 2022.

\bibitem{fsd2023}
Lue Fan, Feng Wang, Naiyan Wang, and Zhaoxiang Zhang.
\newblock Fully sparse {3D} object detection.
\newblock In {\em Advances in Neural Information Processing Systems}. NeurIPS, 2022.

\bibitem{Geiger2012CVPR}
Andreas Geiger, Philip Lenz, and Raquel Urtasun.
\newblock Are we ready for autonomous driving? the {KITTI} vision benchmark suite.
\newblock pages 3354--3361, 2012.

\bibitem{Han_2023}
Yushan Han, Hui Zhang, Huifang Li, Yi~Jin, Congyan Lang, and Yidong Li.
\newblock Collaborative perception in autonomous driving: Methods, datasets, and challenges.
\newblock {\em IEEE Intelligent Transportation Systems Magazine}, 15(6):131--151, 2023.

\bibitem{hu2022where2comm}
Yue Hu, Shaoheng Fang, Zixing Lei, Yiqi Zhong, and Siheng Chen.
\newblock Where2comm: Communication-efficient collaborative perception via spatial confidence maps.
\newblock In {\em Advances in Neural Information Processing Systems}. NeurIPS, 2022.

\bibitem{hu2024codefilling}
Yue Hu, Juntong Peng, Sifei Liu, Junhao Ge, Si~Liu, and Siheng Chen.
\newblock Communication-efficient collaborative perception via information filling with codebook.
\newblock In {\em Proceedings of the IEEE/CVF Conference on Computer Vision and Pattern Recognition}, pages 15481--15490. IEEE, 2024.

\bibitem{bevfusion2024dhip}
Taeho Kim and Joohee Kim.
\newblock {BEVFusion} with dual hard instance probing for multimodal {3D} object detection.
\newblock {\em IEEE Access}, 13, 2025.

\bibitem{kingma2014adam}
Diederik~P. Kingma and Jimmy Ba.
\newblock Adam: A method for stochastic optimization.
\newblock In {\em Proceedings of the International Conference on Learning Representations}. ICLR, 2015.

\bibitem{kong2023dusa}
Xianghao Kong, Wentao Jiang, Jinrang Jia, Yifeng Shi, Runsheng Xu, and Si~Liu.
\newblock {DUSA}: Decoupled unsupervised sim2real adaptation for vehicle-to-everything collaborative perception.
\newblock In {\em Proceedings of the ACM International Conference on Multimedia}, pages 1943--1954. ACM, 2023.

\bibitem{lei2022latencyaware}
Zixing Lei, Shunli Ren, Yue Hu, Wenjun Zhang, and Siheng Chen.
\newblock Latency-aware collaborative perception.
\newblock In {\em Proceedings of the European Conference on Computer Vision}. Springer, 2022.

\bibitem{Li_2021_NeurIPS}
Yiming Li, Shunli Ren, Pengxiang Wu, Siheng Chen, Chen Feng, and Wenjun Zhang.
\newblock Learning distilled collaboration graph for multi-agent perception.
\newblock In {\em Advances in Neural Information Processing Systems}. NeurIPS, 2021.

\bibitem{lin2017focal}
Tsung-Yi Lin, Priya Goyal, Ross Girshick, Kaiming He, and Piotr Doll{\'a}r.
\newblock Focal loss for dense object detection.
\newblock In {\em Proceedings of the IEEE International Conference on Computer Vision}, pages 2980--2988, Venice, Italy, 2017. IEEE.

\bibitem{lu2024heal}
Yifan Lu, Yue Hu, Yiqi Zhong, Dequan Wang, Yanfeng Wang, and Siheng Chen.
\newblock {HEAL}: An extensible framework for open heterogeneous collaborative perception.
\newblock In {\em Proceedings of the International Conference on Learning Representations}. ICLR, 2024.

\bibitem{lu2023robust}
Yifan Lu, Quanhao Li, Baoan Liu, Mehrdad Dianati, Chen Feng, Siheng Chen, and Yanfeng Wang.
\newblock Robust collaborative {3D} object detection in presence of pose errors.
\newblock In {\em Proceedings of the IEEE International Conference on Robotics and Automation}. IEEE, 2023.

\bibitem{NHTSA_2024}
{National Highway Traffic Safety Administration}.
\newblock Overview of motor vehicle traffic crashes in 2022.
\newblock \url{https://crashstats.nhtsa.dot.gov}, 2024.
\newblock Accessed: 2024-10-01.

\bibitem{pang2019libra}
Jiangmiao Pang, Kai Chen, Jianping Shi, Huajun Feng, Wanli Ouyang, and Dahua Lin.
\newblock Libra {R-CNN}: Towards balanced learning for object detection.
\newblock In {\em Proceedings of the IEEE/CVF Conference on Computer Vision and Pattern Recognition}, pages 821--830. IEEE, 2019.

\bibitem{shenkut2025revqom}
Dereje Shenkut and B.~V. K.~Vijaya Kumar.
\newblock {ReVQom}: Residual vector quantization for communication-efficient multi-agent perception.
\newblock {\em arXiv preprint arXiv:2509.21464}, 2025.

\bibitem{shenkut2024impact}
Dereje Shenkut and B.V.K.~Vijaya Kumar.
\newblock Impact of latency and bandwidth limitations on the safety performance of collaborative perception.
\newblock In {\em Proceedings of the International Conference on Computer Communications and Networks}, pages 1--8. IEEE, 2024.

\bibitem{shrivastava2016training}
Abhinav Shrivastava, Abhinav Gupta, and Ross Girshick.
\newblock Training region-based object detectors with online hard example mining.
\newblock In {\em Proceedings of the IEEE Conference on Computer Vision and Pattern Recognition}, pages 761--769, Las Vegas, NV, 2016. IEEE.

\bibitem{song2025trafalign}
Zhiying Song, Lei Yang, Fuxi Wen, and Jun Li.
\newblock {TraF-Align}: Trajectory-aware feature alignment for asynchronous multi-agent perception.
\newblock In {\em Proceedings of the IEEE/CVF Conference on Computer Vision and Pattern Recognition}. IEEE, 2025.

\bibitem{waymo2020}
Pei Sun, Henrik Kretzschmar, Xerxes Dotiwalla, Aurelien Chouard, Vijaysai Patnaik, Paul Tsui, James Guo, Yin Zhou, Yuning Chai, Benjamin Caine, et~al.
\newblock Scalability in perception for autonomous driving: Waymo open dataset.
\newblock In {\em Proceedings of the IEEE/CVF Conference on Computer Vision and Pattern Recognition}, pages 2446--2454, Seattle, WA, 2020. IEEE.

\bibitem{wang2023umc}
Tianhang Wang, Guang Chen, Kai Chen, Zhengfa Liu, Bo~Zhang, Alois Knoll, and Changjun Jiang.
\newblock {UMC}: A unified bandwidth-efficient and multi-resolution based collaborative perception framework.
\newblock In {\em Proceedings of the IEEE/CVF International Conference on Computer Vision}, pages 8187--8196. IEEE, 2023.

\bibitem{wang2020v2vnet}
Tsun-Hsuan Wang, Sivabalan Manivasagam, Ming Liang, Bin Yang, Wenyuan Zeng, and Raquel Urtasun.
\newblock {V2VNet}: Vehicle-to-vehicle communication for joint perception and prediction.
\newblock In {\em Proceedings of the European Conference on Computer Vision}. Springer, 2020.

\bibitem{xia2024hinted}
Qiming Xia, Wei Ye, Hai Wu, Shijia Zhao, Leyuan Xing, Xun Huang, Jinhao Deng, Xin Li, Chenglu Wen, and Cheng Wang.
\newblock {HINTED}: Hard instance enhanced detector with mixed-density feature fusion for sparsely-supervised {3D} object detection.
\newblock In {\em Proceedings of the IEEE/CVF Conference on Computer Vision and Pattern Recognition}, pages 15321--15330. IEEE, 2024.

\bibitem{v2xreal}
Hao Xiang, Zhaoliang Zheng, Xin Xia, Runsheng Xu, Letian Gao, Zewei Zhou, Xu~Han, Xinkai Ji, Mingxi Li, Zonglin Meng, Li~Jin, Mingyue Lei, Zhaoyang Ma, Zihang He, Haoxuan Ma, Yunshuang Yuan, Yingqian Zhao, and Jiaqi Ma.
\newblock {V2X-Real}: A large-scale dataset for vehicle-to-everything cooperative perception.
\newblock In {\em Proceedings of the European Conference on Computer Vision}, pages 455--470. Springer, 2024.

\bibitem{DI_V2X_2023}
Li~Xiang, Junbo Yin, Wei Li, Cheng-Zhong Xu, Ruigang Yang, and Jianbing Shen.
\newblock {DI-V2X}: Learning domain-invariant representation for vehicle-infrastructure collaborative {3D} object detection.
\newblock {\em arXiv preprint arXiv:2312.15742}, 2023.

\bibitem{xu2023bridging}
Runsheng Xu, Jinlong Li, Xiaoyu Dong, Hongkai Yu, and Jiaqi Ma.
\newblock Bridging the domain gap for multi-agent perception.
\newblock In {\em Proceedings of the IEEE International Conference on Robotics and Automation}. IEEE, 2023.

\bibitem{xu2022v2xvit}
Runsheng Xu, Hao Xiang, Zhengzhong Tu, Xin Xia, Ming-Hsuan Yang, and Jiaqi Ma.
\newblock {V2X-ViT}: Vehicle-to-everything cooperative perception with vision transformer.
\newblock In {\em Proceedings of the European Conference on Computer Vision}. Springer, 2022.

\bibitem{xu2022opencood}
Runsheng Xu, Hao Xiang, Xin Xia, Xu~Han, Jinlong Li, and Jiaqi Ma.
\newblock {OPV2V}: An open benchmark dataset and fusion pipeline for perception with vehicle-to-vehicle communication.
\newblock In {\em Proceedings of the IEEE International Conference on Robotics and Automation}. IEEE, 2022.

\bibitem{yan2018second}
Yan Yan, Yuxing Mao, and Bo~Li.
\newblock {SECOND}: Sparsely embedded convolutional detection.
\newblock {\em Sensors}, 18(10), 2018.

\bibitem{yang2023how2comm}
Dingkang Yang, Kun Yang, Yuzheng Wang, Jing Liu, Zhi Xu, Rongbin Yin, Peng Zhai, and Lihua Zhang.
\newblock {How2comm}: Communication-efficient and collaboration-pragmatic multi-agent perception.
\newblock In {\em Advances in Neural Information Processing Systems}. NeurIPS, 2023.

\bibitem{yang2025v2xradar}
Lei Yang, Xinyu Zhang, Jun Li, Chen Wang, Jiaqi Ma, Zhiying Song, Tong Zhao, Ziying Song, Li~Wang, Mo~Zhou, Yang Shen, and Chen Lv.
\newblock {V2X-Radar}: A multi-modal dataset with {4D} radar for cooperative perception.
\newblock In {\em Advances in Neural Information Processing Systems}. NeurIPS, 2025.

\bibitem{yu2022dairv2x}
Haibao Yu, Yizhen Luo, Mao Shu, Yiyi Huo, Zebang Yang, Yifeng Shi, Zhenglong Guo, Hanyu Li, Xing Hu, Jirui Yuan, and Zaiqing Nie.
\newblock {DAIR-V2X}: A large-scale dataset for vehicle-infrastructure cooperative {3D} object detection.
\newblock In {\em Proceedings of the IEEE/CVF Conference on Computer Vision and Pattern Recognition}, pages 21361--21370. IEEE, 2022.

\bibitem{yuan2025sparsealign}
Yunshuang Yuan, Yan Xia, Daniel Cremers, and Monika Sester.
\newblock {SparseAlign}: A fully sparse framework for cooperative object detection.
\newblock In {\em Proceedings of the IEEE/CVF Conference on Computer Vision and Pattern Recognition}. IEEE, 2025.

\bibitem{Zhang_2024_CVPR}
Jingyu Zhang, Kun Yang, Yilei Wang, Hanqi Wang, Peng Sun, and Liang Song.
\newblock Ermvp: Communication-efficient and collaboration-robust multi-vehicle perception in challenging environments.
\newblock In {\em Proceedings of the IEEE/CVF Conference on Computer Vision and Pattern Recognition (CVPR)}, pages 12575--12584, June 2024.

\bibitem{zhou2018voxelnet}
Yin Zhou and Oncel Tuzel.
\newblock {VoxelNet}: End-to-end learning for point cloud based {3D} object detection.
\newblock In {\em Proceedings of the IEEE Conference on Computer Vision and Pattern Recognition}, pages 4490--4499. IEEE, 2018.

\end{thebibliography}

}
\end{document}